\documentclass[runningheads]{llncs}
\usepackage[T1]{fontenc}
\usepackage{graphicx,verbatim}
\usepackage{graphicx}
\usepackage{algorithm}
\usepackage{algpseudocode}
\usepackage{amsmath}
\usepackage{amsfonts}
\usepackage{comment}
\DeclareMathOperator*{\argmin}{arg\,min}
\algnewcommand{\LeftComment}[1]{\Statex \(\triangleright\) #1}
\usepackage{array}
\newcolumntype{P}[1]{>{\centering\arraybackslash}p{#1}}
\usepackage{xcolor}
\usepackage{authblk}
\usepackage{multirow}
\usepackage{multicol}
\makeatletter
\newcommand\footnoteref[1]{\protected@xdef\@thefnmark{\ref{#1}}\@footnotemark}
\makeatother
\begin{document}
\title{Metric-Guided Conformal Bounds for Probabilistic Image Reconstruction}

\author{Matt Y. Cheung\inst{1,2}
\orcidID{0000-0002-7846-3297} 
\and
Tucker J. Netherton\inst{2}
\orcidID{0000-0003-1583-7121} 
\and
Laurence E. Court\inst{2}
\orcidID{0000-0002-3241-6145}
\and
Ashok Veeraraghavan\inst{1}
\orcidID{0000-0001-5043-7460}
\and
Guha Balakrishnan\inst{1}
\orcidID{0000-0001-8703-1368}
}
\authorrunning{Cheung et al.}

\institute{Department of Electrical \& Computer Engineering, Rice University, Houston TX, USA \and
Department of Radiation Physics, The University of Texas M.D. Anderson Cancer Center, Houston TX, USA
\\ \email{guha@rice.edu}}

\authorrunning{M. Cheung et al.}

\maketitle              
\begin{abstract}
    Modern deep learning reconstruction algorithms generate impressively realistic scans from sparse inputs, but can often produce significant inaccuracies. This makes it difficult to provide statistically guaranteed claims about the true state of a subject from scans reconstructed by these algorithms. In this study, we propose a framework for computing provably valid prediction bounds on claims derived from probabilistic black-box image reconstruction algorithms. The key insights are to represent reconstructed scans with a derived clinical metric of interest and to calibrate bounds on the ground truth metric with conformal prediction (CP) using a prior calibration dataset. These bounds convey interpretable feedback about the subject's state, and can also be used to retrieve nearest-neighbor reconstructed scans for visual inspection. We demonstrate the utility of this framework on sparse-view computed tomography (CT) for fat mass quantification and radiotherapy planning tasks. Results show that our framework produces visual bounds with better semantical interpretation than conventional pixel-based bounding approaches and captures important spatial correlations. Furthermore, we can flag dangerous outlier reconstructions that look plausible but have statistically unlikely metric values. Code available at: \url{https://github.com/matthewyccheung/conformal-metric} 
\end{abstract}
\section{Introduction}
\label{sec:introduction}
Classical sparse medical imaging reconstruction algorithms -- from Iterative Reconstruction (IR) methods~\cite{herman2009fundamentals} used in computed tomography (CT) to compressive sensing methods~\cite{lustig2008compressed} used in magnetic resonance imaging (MRI) -- have one extremely desirable property: when they fail, they \emph{\textbf{fail}}. That is, in most cases, ``incorrect'' reconstructions also look obviously poor to the naked eye, and we understand their modes of failure. In contrast, modern deep generative and neural rendering algorithms~\cite{jalal2021robust,lakshminarayanan2017simple,zha2022naf} use rich implicit or data-driven priors to produce impressive looking images, but are known to \emph{hallucinate}, i.e., make predictions which are inaccurate but look plausible~\cite{bhadra2021hallucinations,gawlikowski2023survey,nguyen2015deep}. This creates an operational challenge in engendering trust for these algorithms in medical imaging. 

In the typical image reconstruction pipeline, variations in observation acquisition (e.g., patient movement, sensor noise)~\cite{diwakar2018review,kalisz2016artifacts} and lossiness of observations contribute to reconstruction uncertainty. Deep learning-based reconstruction algorithms can account for some of this uncertainty by randomizing inputs or parameters to produce a distribution of solutions, but these approaches are not guaranteed to capture the full space of solutions. In particular, \emph{epistemic uncertainty} caused by missing data in the training distribution is likely to be underestimated and contribute to hallucinations~\cite{chan2024estimating}.

\begin{figure*}[t!]
    \centering
    \includegraphics[width=\linewidth]{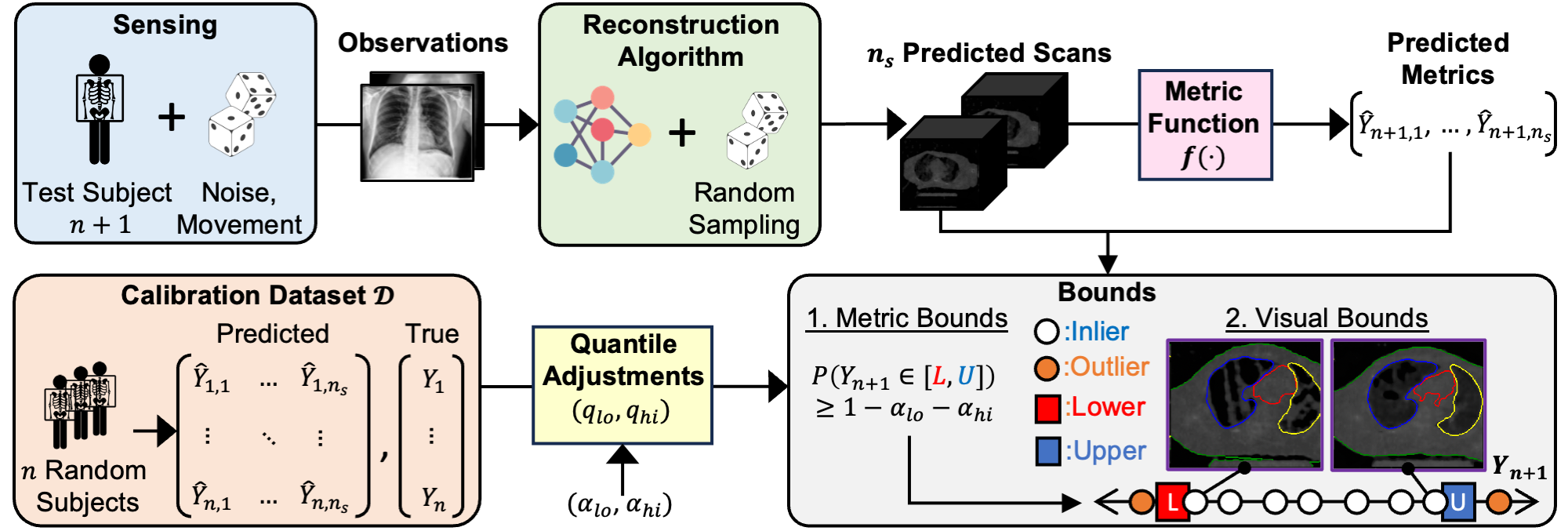}
    \caption{\textbf{Overview of our approach.} Standard imaging pipeline with probabilistic image acquisition, probabilistic image reconstruction algorithm, reconstructed volumes, and a downstream application. We propose to use CP on downstream metrics to attain meaningful lower and upper bounds, inliers, and outliers. Lower and upper bound images correspond to the reconstructions with metrics closest to the CP prediction interval lower and upper bounds (nearest neighbors or NN).
    CP prediction intervals are constructed based on a separate calibration set.}
    \label{fig:overview}
\end{figure*}

This raises a crucial question: For a given test subject, given a random sample of reconstructed scans produced by a black-box deep learning algorithm, is it possible to make statistically guaranteed claims about the subject's true state? We show that this is possible and propose a framework based on two ideas (Fig.~\ref{fig:overview}). First, we make such statistical claims over a meaningful downstream metric derived from the scans, e.g., heart volume or fat content, rather than over raw pixel values, which suffer from high dimensionality and limited interpretability. Second, we assume an available \emph{calibration dataset} consisting of metric values derived from prior reconstructed and ground-truth scans. At test time, this calibration dataset may be used to adjust for systematic deviations of metrics derived from the reconstructions with respect to ground-truth scans.

More specifically, we first algorithmically compute a metric of interest from each sample in a set of $n_s$ reconstructed scans for a test subject (Fig.~\ref{fig:overview}-top). Second, using this set of metric values, we compute valid prediction intervals that reliably contain the ground truth value on average using a calibration dataset $\mathcal{D}$ of $n$ random subjects (Fig.~\ref{fig:overview}-bottom). To do so, we build on conformal prediction (CP), a powerful technique to construct uncertainty-aware prediction intervals for values predicted by a black-box algorithm~\cite{fontana2023conformal,papadopoulos2002inductive,shafer2008tutorial,vovk2005algorithmic}. Under the assumption of exchangeability, the resulting bounds will provably contain the ground truth values $(1-\alpha_{lo}-\alpha_{hi})\%$ of the time for some set lower/upper mis-coverage rates $\alpha_{lo}/\alpha_{hi}$. Finally, we can link the lower/upper metric bounds to their nearest reconstructed scans in the sample set to provide visual interpretation. 

\begin{algorithm}[t!]
    \caption{Metric-Guided Bound Computation.}
    \begin{algorithmic}
        \Statex \textbf{Inputs: } Calibration set $\mathcal{D} = \{ \hat Y_i \}_{i=1}^n$, test subject sample metrics $\hat{Y}_{n+1}$, lower and upper miscoverage rates $\alpha_{lo}$ and $\alpha_{hi}$.
        \Statex \textbf{Outputs: } Prediction interval $C(\hat Y_{n+1})$, lower and upper bound reconstructions $L(\hat I_{n+1})$ and $U(\hat I_{n+1})$, inlier and outlier reconstructions $In(\hat I_{n+1})$ and $Out(\hat I_{n+1})$.
        \Statex \LeftComment{Perform calibration to get quantile adjustments}
        \For{$i=1:n$}
            \State $[s_{i,lo},s_{i,hi}] \leftarrow [Q_{\alpha_{lo}}(\hat Y_i)-Y_i, Y_i-Q_{1-\alpha_{hi}}(\hat Y_i)]$
        \EndFor
        \State $\hat \alpha_{lo}, \hat \alpha_{hi} \leftarrow \lfloor\alpha_{lo}(n+1)\rfloor/n, \lfloor\alpha_{hi}(n+1)\rfloor/n$ 
        \State $[q_{lo},q_{hi}] \leftarrow [Q_{1-\hat \alpha_{lo}}(\{s_{i,lo}\}_{i=1}^n), Q_{1-\hat \alpha_{hi}}(\{s_{i,hi}\}_{i=1}^n)]$
        \State \LeftComment{Compute prediction interval for test patient.}
        \State $C(\hat Y_{n+1})\leftarrow [Q_{\alpha_{lo}}(\hat Y_{n+1}) - q_{lo}, Q_{1-\alpha_{hi}}(\hat Y_{n+1}) + q_{hi}]$
    \end{algorithmic}
    \label{alg:metricbounds}
\end{algorithm}

Most existing strategies applying CP to image reconstruction operate directly on pixel values~\cite{angelopoulos2022image,horwitz2022conffusion,teneggi2023trust}, which yield bounds with limited interpretability and do not capture spatial correlations. Other studies propose using low-dimensional representations of pixels computed using principal component analysis (PCA)~\cite{belhasin2023principal} or latent spaces of pretrained generative models~\cite{sankaranarayanan2022semantic}. However, PCA is expensive to compute for scans larger than a small size, and high-quality generative models are not available for arbitrary medical image sets. In contrast, we propose computing CP bounds over metrics derived from the reconstructed scans, resulting in a flexible, interpretable, and computationally tractable solution.

We demonstrate the utility of our method on sparse-view CT reconstruction. We focus specifically on the downstream metrics for fat mass quantification and radiotherapy (RT) planning using de-identified CT datasets from The University of Texas M.D. Anderson Cancer Center.
Our results first show that our method achieves valid coverage for downstream metrics while more common pixel-based bounding methods exhibit significant undercoverage, i.e., generated lower and upper bounds do not contain the ground truth metric with a user-specified probability. Second, we demonstrate that meaningful visual bounds in the form of nearest neighbor reconstructed scans can be reliably retrieved from the metric bounds. Finally, we show that our framework flags dangerous outlier reconstructions that look plausible but have statistically unlikely metric values. Our work lays the foundation for more interpretable and trustworthy test-time assessments of medical image reconstruction algorithms.

\section{Methods}
\label{sec:method}
We assume a fixed stochastic image acquisition system and reconstruction algorithm that outputs scans in $\mathbb{R}^\Omega$, and a given function $f(\cdot): \mathbb{R}^\Omega \rightarrow \mathbb{R}^1$ that outputs a scalar metric of interest from a scan. We also assume an input calibration dataset $\mathcal{D} = \{ \hat Y_i \}_{i=1}^n$ of $n$ random subjects, where $\hat Y_i = \{\hat Y_{i,j}\}_{j=1}^{n_s}$ is a set of $n_s$ predicted metrics (derived by running $f(\cdot)$ on $n_s$ randomly sampled reconstructed scans). Let $\alpha_{lo}$ and $\alpha_{hi}$ denote desired lower and upper mis-coverage rates, and $Q_{\alpha}(\cdot)$ a function that returns the $\alpha$-th quantile of a set.

For test subject $n+1$, we derive a prediction set $C(\hat Y_{n+1})=[L(\hat Y_{n+1}),U(\hat Y_{n+1})]$ that yields marginal coverage: $P(Y_{n+1}\in C(\hat{Y}_{n+1}))\geq 1-\alpha_{lo}-\alpha_{hi}$~\cite{fontana2023conformal}.
If the reconstruction pipeline were perfectly calibrated, we could simply set $L(\hat Y_{n+1})=Q_{\alpha_{lo}}(\hat Y_{n+1})$ and $U(\hat Y_{n+1})=Q_{\alpha_{hi}}(\hat Y_{n+1})$. However,to handle mis-calibrations, we use $\mathcal{D}$ to find appropriate adjustments to $Q_{\alpha_{lo}}(\hat Y_{n+1})$ and $Q_{\alpha_{hi}}(\hat Y_{n+1})$. We find these adjustments in Sec.~\ref{sec:adjustment}, and the use of the metric bounds to retrieve inlier, outlier, and nearest neighbor reconstructed scans in Sec.~\ref{sec:visualbounds}.



\begin{algorithm}[t!]
    \caption{Metric-Guided Image Bound Retrieval and Outlier Detection.}
    \begin{algorithmic}
        \Statex \textbf{Inputs: } Test subject sample images $\hat{I}_{n+1}$ and metrics $\hat{Y}_{n+1}$, Prediction interval $C(\hat Y_{n+1})=[L(\hat{Y}_{n+1}),U(\hat{Y}_{n+1})]$ from Alg. \ref{alg:metricbounds}.
        \Statex \textbf{Outputs: } Lower and upper bound reconstructions $L(\hat I_{n+1})$ and $U(\hat I_{n+1})$, inlier and outlier reconstructions $In(\hat I_{n+1})$ and $Out(\hat I_{n+1})$.

        \State \LeftComment{Retrieve upper and lower bound reconstructions.}
        \State $L(\hat I_{n+1}) \leftarrow \argmin_{\hat I_{n+1,j}}{|\hat Y_{n+1,j} - L(\hat Y_{n+1})|}$
        \State $U(\hat I_{n+1}) \leftarrow \argmin_{\hat I_{n+1,j}}{|\hat Y_{n+1,j} - U(\hat Y_{n+1})|}$
        \State \LeftComment{Retrieve inliers and outlier reconstructions.}
        \State $In(\hat I_{n+1}) \leftarrow \{\hat I_{n+1,j} \in \hat I_{n+1} | \hat Y_{n+1,j} \in [L(\hat Y_{n+1}), U(\hat Y_{n+1})]\}$
        \State $Out(\hat I_{n+1}) \leftarrow \{\hat I_{n+1,j} \in \hat I_{n+1} | \hat Y_{n+1,j} \notin [L(\hat Y_{n+1}), U(\hat Y_{n+1})]\}$
    \end{algorithmic}
    \label{alg:imageretrieval}
\end{algorithm}
\subsection{Quantile Adjustments Using Calibration Data}
\label{sec:adjustment}
Using $\mathcal{D}$, we derive adjustments $q_{lo}$ and $q_{hi}$ to the $Q_{\alpha_{lo}}(\hat{Y}_{n+1})$ and $Q_{1-\alpha_{hi}}(\hat{Y}_{n+1})$ quantiles of the ground truth metric of test subject $n+1$ (Alg.~\ref{alg:metricbounds}). To do so, we use a probabilistic ``split'' CP setup proposed in previous work~\cite{wang2022probabilistic}. We first calculate $s_{lo,i}$ and $s_{hi,i}$, the lower and upper \emph{non-conformity scores} for each patient, capturing the differences between the lower and upper sample quantiles and ground truth metrics: $s_{i,lo}=Q_{\alpha_{lo}}(\hat Y_i)-Y_i$ and $s_{i,hi}=Y_i-Q_{1-\alpha_{hi}}(\hat Y_i)$. We then compute the $(1-\alpha_{lo})$-th and $(1-\alpha_{hi})$-th empirical quantiles of these scores to arrive at the adjustments: $q_{lo}=Q_{1-\hat \alpha_{lo}}(\{s_{i,lo}\}_{i=1}^n)$ and $q_{hi}=Q_{1-\hat \alpha_{hi}}(\{s_{i,hi}\}_{i=1}^n)$, where $\hat \alpha_{lo}=\lfloor\alpha_{lo}(n+1)\rfloor/n$ and $\hat \alpha_{hi}=\lfloor\alpha_{hi}(n+1)\rfloor/n$ are finite-sample adjusted mis-coverage rates. 

At test time, we first reconstruct images $\hat I_{n+1}=\{\hat I_{n+1,j}\}_{j=1}^{n_s}$ for the test subject and compute the derived metrics $\hat Y_{n+1} = \{\hat Y_{n+1,j}\}_{j=1}^{n_s}$. We then simply compute the prediction interval $C(\hat Y_{n+1})=[Q_{\alpha_{lo}}(\hat Y_{n+1})-q_{lo},Q_{1-\alpha_{hi}}(\hat Y_{n+1})+q_{hi}]$ using the learned adjustments from $\mathcal{D}$. Under the assumption of exchangeability, this prediction interval is guaranteed to have marginal coverage~\cite{wang2022probabilistic}.

\subsection{Retrieving Scans Using Metric Bounds}
\label{sec:visualbounds}
While the metric bounds are useful in their own right to offer interpretable feedback on patient state, they can also be mapped back to the reconstructed scans $\{\hat Y_{n+1,j}\}_{j=1}^{n_s}$ to provide visual feedback to the user or clinician (Alg.~\ref{alg:imageretrieval}). First, we can flag ``outlier'' scans, i.e., those scans with metrics outside $[L(\hat I_{n+1}), U(\hat I_{n+1})]$. Second, we can retrieve the scans with metrics nearest to $L(\hat I_{n+1})$ and $U(\hat I_{n+1})$, thereby providing visual interpretability to the metric bounds.

\begin{figure*}[t!]
    \centering    
    \includegraphics[width=\linewidth]{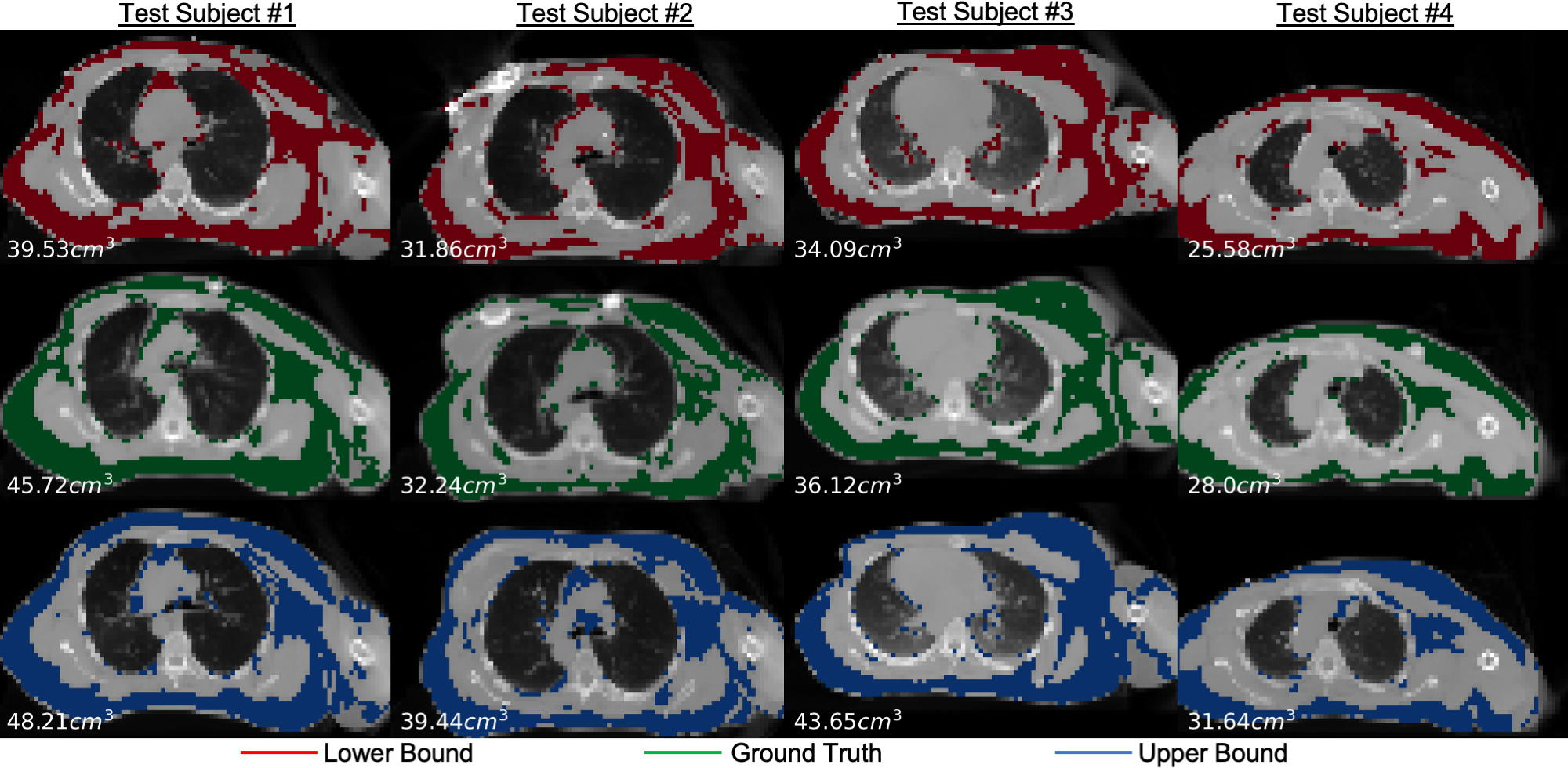}
    \caption{\textbf{Sample retrieved visual bounds for fat mass quantification.}
    For four subjects, we show retrieved reconstructions closest to the metric lower bound (top row) and upper bound (bottom), along with the ground truth scan (middle). Per scan, we overlay predicted fat pixels and print total fat volume on the bottom left. The visual bounds provide context into which spatial regions significantly vary in the prediction interval, which may not be obvious from the metric bounds alone.}
    \label{fig:fat_bounds}
\end{figure*}

\section{Experiments}
We empirically evaluated our framework on sparse-view CT reconstruction applications: thoracic fat mass quantification (fat) and post-mastectomy radiotherapy planning (RT), described below. We compared our approach, named \emph{Metric CP}, to three baselines: \emph{Metric}, \emph{Pixel}, and \emph{Pixel CP}. \emph{Metric} simply computes the $\alpha_{lo}$-th and $(1-\alpha_{hi})$-th quantiles of metrics across sampled images. \emph{Pixel} computes pixel-wise $\alpha_{lo}$-th and $(1-\alpha_{hi})$-th quantiles of intensity values across images~\cite{edupuganti2020uncertainty,gillmann2021uncertainty,jalal2021robust}. \emph{Pixel CP} applies CP to adjust the pixel-wise quantiles to achieve marginal coverage of ground-truth pixel intensities~\cite{romano2019conformalized,wang2022probabilistic}. For all experiments, we set $\alpha_{lo}=\alpha_{hi}=0.05$ (target coverage of 0.9).

\textbf{Fat Mass Estimation.} Quantifying thoracic fat mass distribution is crucial to understanding its role in cardiometabolic risk and other health outcomes~\cite{dey2012epicardial,wang2014imaging}. We computed fat volume for a 2D CT slice by multiplying the number of fat pixels with slice thickness. We identified fat pixels by thresholding values in [-150,-50] Hu. We used a de-identified CT dataset of 935 subjects from The University of Texas M.D. Anderson Cancer Center.
\footnote{\label{note1}This research was conducted using an approved institutional review board protocol.}.
We split this dataset into 729 subjects ($\sim$300k axial slices) for training and 265 subjects (795 axial slices) for testing. We trained an unconditional 2D U-Net diffusion model on training data and used guided posterior sampling to reconstruct slices from 15 1-D observed projections~\cite{dhariwal2021diffusion,graikos2022diffusion}. We used $n=596$ slices for calibration and 199 slices for testing, and generated $n_s=100$ reconstructions per slice using random input noise initializations. We repeated the experiment 1000 times with random calibration-testing splits. 

\begin{figure*}[t!]
    \centering    
    \includegraphics[width=\linewidth]{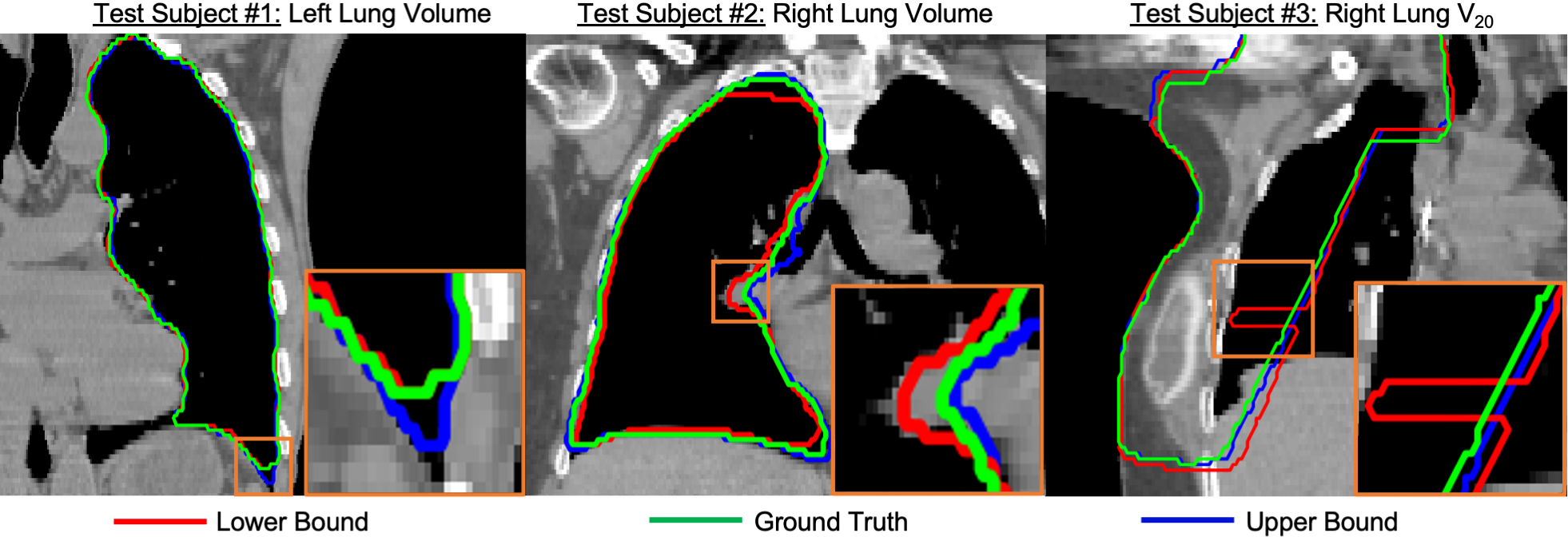}
    \caption{\textbf{Sample retrieved visual bounds for RT planning.} We overlay contours on ground truth CT scans corresponding to RT metrics derived from: the ground truth scan (green), and reconstructed scans closest to the metric lower bound (red), and upper bound (blue). We show a different RT planning metric per subject. The bottom right inset of each scan zooms in on one region with significant contour differences. Left/right lung volume metrics are sensitive to reconstruction quality at small ``corner''-like regions. The lower bound for $V_{20}$ (subject 3) contains a significant spike, indicating sensitivity to some reconstructed artifact in that spatial region.}
    \label{fig:dose_bounds}
\end{figure*}

\textbf{RT Planning.} RT is a cornerstone of modern cancer treatment, and sparse-view CT can potentially lower equipment costs and improve access to CT imaging for RT planning~\cite{court2023addressing}. We generated RT dose plans for CTs using the Radiation Planning Assistant (RPA, FDA 510(k) cleared)~\cite{aggarwal2023radiation,court2023addressing,kisling2018radiation} and investigated several key dose and structural metrics: maximum dose to the heart/left lung/right lung ($D_0$), volumes of heart/left lung/right lung (Vol), and volume of right lung that received 20Gy (Right Lung $V_{20}$). We used a de-identified CT dataset of 40 subjects retrospectively treated with radiotherapy at The University of Texas M.D. Anderson Cancer Center. For each subject, we generated digitally reconstructed radiographs (DRRs) from the ground truth cone-beam CT at 50 uniformly spaced angles between 0$^{\circ}$ and 360$^{\circ}$. 
We used a self-supervised reconstruction model, Neural Attenuation Fields (NAF)~\cite{zha2022naf}. We used $n=30$ scans for calibration and 10 scans for testing, and generated $n_s=10$ reconstructions per scan using random initializations of network parameters. We repeated the experiment 1000 times with random calibration-testing splits.

\begin{table}[t]
\centering
\caption{\textbf{Quantitative comparison of \emph{Metric CP} to baselines.} We used 1000 random $75\%$-$25\%$ calibration-testing splits, and report mean coverage and normalized interval lengths. The normalized interval length for a baseline method is its mean interval length divided by \emph{Metric CP}'s mean interval length. While \emph{Metric}, \emph{Pixel}, and \emph{Pixel CP} bounds often have smaller (``tighter'') interval lengths than \emph{Metric CP} (a normalized length $<1$), they suffer from significant under-coverage, making them unusable for interpretable confidence assessments. By construction, \emph{Metric CP} provides a target coverage with reasonable interval lengths.}
\resizebox{\columnwidth}{!}{%
\begin{tabular}{cc|cccc|ccc|}
\cline{3-9}
\multicolumn{1}{l}{} &
  \multicolumn{1}{l|}{} &
  \multicolumn{4}{c|}{Test Coverage ($\uparrow$)} &
  \multicolumn{3}{c|}{Normalized Interval Length ($\downarrow$)} \\ \hline
\multicolumn{1}{|c|}{ } &
  Metric &
  \multicolumn{1}{c|}{Metric CP} &
  \multicolumn{1}{c|}{Metric} &
  \multicolumn{1}{c|}{Pixel CP} &
  Pixel &
  \multicolumn{1}{c|}{Metric} &
  \multicolumn{1}{c|}{Pixel CP} &
  Pixel \\ \hline
\multicolumn{1}{|c|}{Fat} &
  Fat Mass &
  \multicolumn{1}{c|}{\textbf{0.902}} &
  \multicolumn{1}{c|}{0.804} &
  \multicolumn{1}{c|}{0.316} &
  0.316 &
  \multicolumn{1}{c|}{0.92} &
  \multicolumn{1}{c|}{3.12} &
  3.12 \\ \hline
\multicolumn{1}{|c|}{\multirow{9}{*}{RT}} &
  Heart $D_0$ &
  \multicolumn{1}{c|}{\textbf{0.932}} &
  \multicolumn{1}{c|}{0.576} &
  \multicolumn{1}{c|}{0.650} &
  0.725 &
  \multicolumn{1}{c|}{0.64} &
  \multicolumn{1}{c|}{0.28} &
  0.24 \\ \cline{2-9} 
\multicolumn{1}{|c|}{} &
  Heart Vol &
  \multicolumn{1}{c|}{\textbf{0.935}} &
  \multicolumn{1}{c|}{0.441} &
  \multicolumn{1}{c|}{0.225} &
  0.475 &
  \multicolumn{1}{c|}{0.23} &
  \multicolumn{1}{c|}{0.19} &
  0.19 \\ \cline{2-9} 
\multicolumn{1}{|c|}{} &
  Right Lung $D_0$ &
  \multicolumn{1}{c|}{\textbf{0.941}} &
  \multicolumn{1}{c|}{0.695} &
  \multicolumn{1}{c|}{0.510} &
  0.418 &
  \multicolumn{1}{c|}{0.66} &
  \multicolumn{1}{c|}{0.16} &
  0.13 \\ \cline{2-9} 
\multicolumn{1}{|c|}{} &
  Right Lung Vol &
  \multicolumn{1}{c|}{\textbf{0.930}} &
  \multicolumn{1}{c|}{0.176} &
  \multicolumn{1}{c|}{0.000} &
  0.000 &
  \multicolumn{1}{c|}{0.72} &
  \multicolumn{1}{c|}{0.00} &
  0.00 \\ \cline{2-9} 
\multicolumn{1}{|c|}{} &
  Right Lung $V_{20}$ &
  \multicolumn{1}{c|}{\textbf{0.934}} &
  \multicolumn{1}{c|}{0.649} &
  \multicolumn{1}{c|}{0.018} &
  0.018 &
  \multicolumn{1}{c|}{0.82} &
  \multicolumn{1}{c|}{0.31} &
  0.31 \\ \cline{2-9} 
\multicolumn{1}{|c|}{} &
  Left Lung $D_0$ &
  \multicolumn{1}{c|}{\textbf{0.958}} &
  \multicolumn{1}{c|}{0.721} &
  \multicolumn{1}{c|}{0.225} &
  0.275 &
  \multicolumn{1}{c|}{0.41} &
  \multicolumn{1}{c|}{0.01} &
  0.01 \\ \cline{2-9} 
\multicolumn{1}{|c|}{} &
  Left Lung Vol &
  \multicolumn{1}{c|}{\textbf{0.935}} &
  \multicolumn{1}{c|}{0.247} &
  \multicolumn{1}{c|}{0.025} &
  0.025 &
  \multicolumn{1}{c|}{0.61} &
  \multicolumn{1}{c|}{0.02} &
  0.02 \\ \hline

\end{tabular}
}
\label{tab1}
\end{table}

\textbf{Evaluation Metrics.} Statistical prediction methods are faced with a trade-off between \emph{test coverage} and \emph{interval length} metrics. Test coverage is the probability that the ground truth metric for a test subject falls within the calibrated prediction interval. Interval length is the difference between upper and lower bounds (and 0 if the difference is negative). For baseline methods, we specifically report normalized interval length: interval lengths divided by \emph{Metric CP}'s interval length, thereby removing arbitrary scaling differences across metrics. 


\subsection{Results}
Table~\ref{tab1} presents quantitative results. \emph{Metric CP} achieves valid coverage for each metric of choice, while other common baseline strategies suffer from significant under-coverage. 
Bounds derived from simply taking 0.05-th and 0.95-th quantiles of sampled metrics (without CP) or pixels therefore do not align well with calibrated metric prediction interval bounds. \emph{Pixel} and \emph{Pixel CP} yield small interval lengths, which may seem beneficial. However, they lead to poor coverage because they focus on pixel intensity, which is often not a key factor relevant to downstream metrics. \emph{Metric} (without CP) still significantly outperforms both pixel approaches, showing that if a user wants a prediction interval for a given metric, it is better to simply estimate this interval using the metric rather than use any pixel-wise approach even with CP. 

\begin{figure*}[t!]
    \centering    \includegraphics[width=\linewidth]{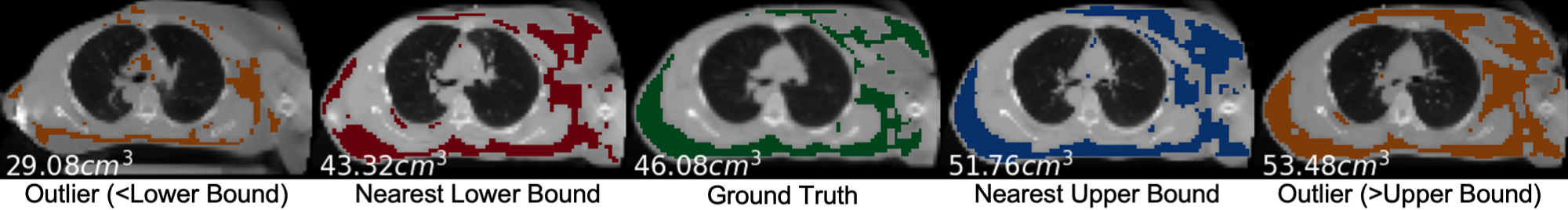}
    \caption{\textbf{Sample retrieved fat mass outliers for one subject.} We show example outlier scans that have metrics outside the calibrated prediction intervals (columns 1 and 5), nearest neighbor lower and upper bound scans (columns 2 and 4), and the ground truth scan (column 3). Although outlier scans may look plausible, our framework flags these scans as not being statistically probable due to their metric values.}
    \label{fig:outliers}
\end{figure*}

Fig.~\ref{fig:fat_bounds} and Fig.~\ref{fig:dose_bounds} present sample visual bound retrievals for fat estimation and RT planning metrics. We overlay per-pixel fat classification predictions on the scans in Fig.~\ref{fig:fat_bounds}, and overlay RT planning contours in Fig.~\ref{fig:dose_bounds}. The visual bounds for fat estimation in Fig.~\ref{fig:fat_bounds} provide context into which spatial regions vary significantly in fat content across the prediction interval. The visual bounds for RT planning in Fig.~\ref{fig:dose_bounds} demonstrate that often there are small regions of an organ that are responsible for the variance to RT planning values. Finally, Fig.~\ref{fig:outliers} shows an example of detecting outliers from reconstructions for one subject for the fat estimation task. Although the outliers in this figure may look plausible, our framework flags these scans as not being statistically probable due to their extreme metric values. It is likely that the diffusion reconstruction algorithm may have some biases contributing to those cases.
\section{Conclusion}
We proposed a theoretically grounded, application-tuned framework to identify the quality of algorithmically reconstructed scans, and derive valid prediction intervals on derived metrics. Results show that representing scans with metrics, as opposed to operating in the raw pixel space, helps produce interpretable confidence intervals that capture meaningful spatial correlations between pixels~\cite{kisling2018radiation,aggarwal2023radiation,court2023addressing}. However, naively computing confidence intervals directly from these metrics without \emph{calibration} is insufficient, as they do not accurately reflect the true distribution of ground truth values. The proposed framework could prove useful in flagging situations where a scan-derived metric is statistically likely to be outside of acceptable safety bounds. For example, maximum dose to the heart in RT planning should be $<$45Gy. If bounds for a subject indicate a good chance of the metric going beyond this limit, the case can be flagged for further review and intervention~\cite{lekeufack2023conformal,romano2020malice,ye2023learned}.

Our framework has a few assumptions and limitations to consider in practice. First, CP assumes exchangeability of data samples. 
Practitioners should ensure that calibration data is representative of future test data. 
Furthermore, the current method focuses on computing bounds on one metric, but there are often several metrics of interest in any application. A natural extension of this work would be to derive multi-metric bounds for image reconstruction. Moreover, image bound retrieval is only useful if there are reconstructions in the scan set with metric values sufficiently close to the metric bounds. If, for example, the reconstruction algorithm has significant bias or the number of samples $n_s$ is small, this may not hold and lead to misleading visual interpretations. Finally, more case studies are warranted to demonstrate the utility of these visual bounds.




\begin{credits}
\subsubsection{\ackname} 
The authors would also like to acknowledge support from a fellowship from the Gulf Coast Consortia on the NLM Training Program in Biomedical Informatics and Data Science T15LM007093. The authors would also like to thank the RPA team (Joy Zhang, Raphael Douglas) for their support. Tucker Netherton would like to acknowledge the support of the NIH LRP award.
\subsubsection{\discintname}
The authors have no competing interests to declare that are relevant to the content of this article.
\end{credits}

\bibliographystyle{splncs04}  
\bibliography{mybibliography}

\end{document}